%% file: full-paper-template.tex
\documentclass[pmlr]{jmlr}

\usepackage[load-configurations=version-1]{siunitx} %

\input{math_commands.tex}

\usepackage{microtype}
\usepackage{subfigure}
\usepackage{wrapfig}
\RequirePackage{graphicx}
 \usepackage{booktabs}
\usepackage{longtable}
\usepackage{hyperref}
\usepackage{algorithm}
\usepackage{algorithmic}
\usepackage{mathtools}
\usepackage{epstopdf}    
\usepackage{amssymb}
\usepackage{pifont}
\newcommand{\cmark}{\ding{51}}%
\newcommand{\xmark}{\ding{55}}%
\usepackage{colortbl} 
\definecolor{myyellow}{RGB}{255, 102, 102}
\definecolor{darkred}{rgb}{0.55, 0.0, 0.0}

\newcommand{\model}{BiTimelyGPT\xspace}
\def\code#1{\texttt{#1}}

\makeatletter
\def\set@curr@file#1{\def\@curr@file{#1}} 
\makeatother
\usepackage[load-configurations=version-1]{siunitx} 

\usepackage{amssymb}
\usepackage{mathtools}
\usepackage{caption}
\usepackage[capitalize,noabbrev]{cleveref}
\usepackage{bm}

\theorembodyfont{\upshape}
\theoremheaderfont{\scshape}
\theorempostheader{:}
\theoremsep{\newline}

\jmlrvolume{252}
\jmlryear{2024}
\jmlrworkshop{Machine Learning for Healthcare}

\title[BiTimelyGPT]{Bidirectional Generative Pre-training for Improving Healthcare Time-series Representation Learning}

\newcommand{\addrcs}{School of Computer Science, McGill University, Montreal, Quebec, Canada}
\newcommand{\addrpop}{School of Population and Global Health, McGill University, Montreal, Quebec, Canada}
\newcommand{\addrmila}{Mila - Quebec AI Institute, Montreal, Quebec, Canada}

\author{\Name{Ziyang Song}$^{1,3}$         \Email{ziyang.song@mail.mcgill.ca}
        \AND
        \Name{Qincheng Lu}$^1$ \Email{qincheng.lu@mail.mcgill.ca}
        \AND
        \Name{He Zhu}$^{1,3}$ \Email{he.zhu2@mail.mcgill.ca}
        \AND
        \Name{David Buckeridge}$^{2}$ \Email{david.buckeridge@mcgill.ca}
        \AND
        \Name{Yue Li}$^{1, 3}$ \Email{yueli@cs.mcgill.ca}
        \AND
        $^1$ \addr \addrcs \\ 
        $^2$ \addr \addrpop \\
        $^3$ \addr \addrmila        
}

\begin{document}

\maketitle

\begin{abstract}
    Learning time-series representations for discriminative tasks, such as classification and regression, has been a long-standing challenge in the healthcare domain. Current pre-training methods are limited in either unidirectional next-token prediction or randomly masked token prediction. We propose a novel architecture called \textbf{Bidirectional Timely Generative Pre-trained Transformer (\model)}\footnote{We have open-sourced source code at \url{https://github.com/li-lab-mcgill/BiTimelyGPT}}, which pre-trains on biosignals and longitudinal clinical records by both next-token and previous-token prediction in alternating transformer layers. This pre-training task preserves original distribution and data shapes of the time-series. Additionally, the full-rank forward and backward attention matrices exhibit more expressive representation capabilities. Using biosignals and longitudinal clinical records, \model demonstrates superior performance in predicting neurological functionality, disease diagnosis, and physiological signs. By visualizing the attention heatmap, we observe that the pre-trained \model can identify discriminative segments from biosignal time-series sequences, even more so after fine-tuning on the task.
\end{abstract}

\section{Introduction}
\label{intro}

Healthcare time-series data can provide valuable insights into patient health status and their clinical outcomes \citep{ma2023survey, TS-TCC}. In the field of healthcare, there are two primary categories: continuous and irregularly sampled time-series data. Continuous time-series, such as biosignals, has been extensively utilized for diverse applications such as disease diagnosis \citep{XSleepNet}, neurophysiological functionality \citep{biosignal_forecasting}, and vital signal estimation \citep{DeepPPG}. Irregularly sampled time series is commonly found in electronic health records (EHRs), where spontaneous updates are made due to outpatient visits or inpatient stays \citep{RAINDROP}. One valuable application is learning temporal representations from healthcare data to enhance sequence-to-vector (seq2vec) prediction tasks. A long-standing challenge lies in learning meaningful representations from the time-series, particularly in scenarios with limited labeled data. To address this problem, self-supervised learning designs pre-training tasks to learn representations from unlabeled time-series data, thereby improving performance in downstream tasks \citep{ma2023survey}.

\begin{figure}[t]
\begin{center}
\centerline{\includegraphics[width=\columnwidth]{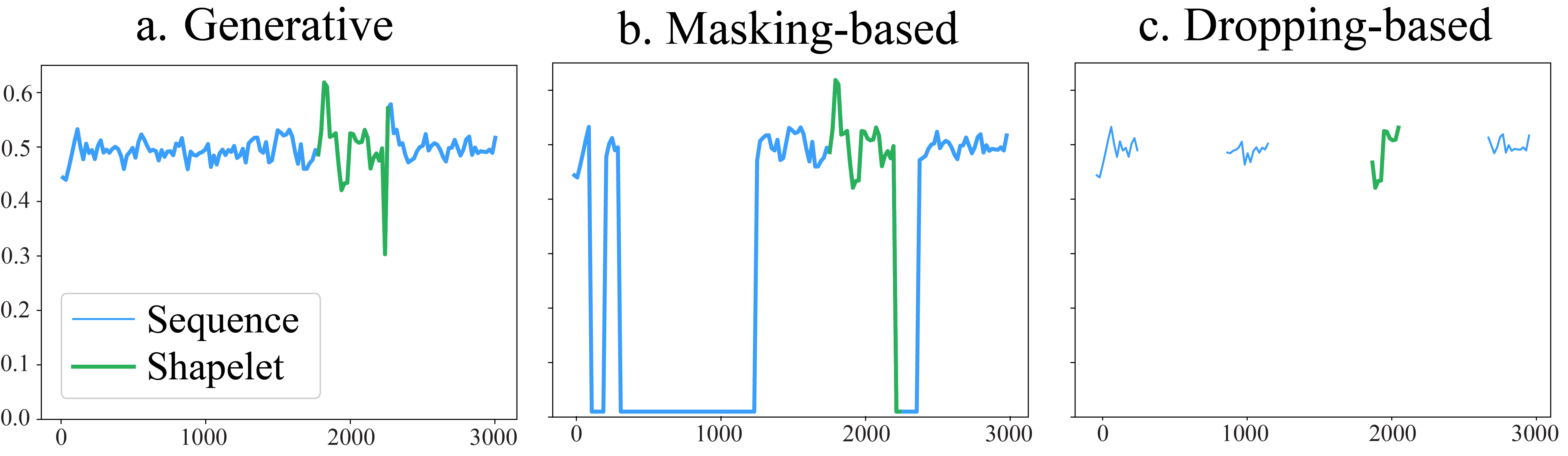}}
\caption{Three pre-training methods on a normalized biosignal sequence. The green segment denotes a discriminative subsequence (shapelet) learned by a published pyts package  \citep{pyts}. \textbf{a.}  \textbf{Generative} pre-training preserves its original distribution and time-series shapelet, maintaining the integrity of the sequence information. \textbf{b.}  \textbf{Masking-based} pre-training masks segments as zeros, resulting in an issue of distribution shift.  \textbf{c.} \textbf{Dropping-based} pre-training randomly discards up to 70\% of the segments and thus disrupts its discriminative shapelet. 
}
\label{pretraining_tasks}
\end{center}
\vskip -0.4in
\end{figure}

Recent advancements in Transformer pre-trained models (PTMs) for Natural Language Processing and Computer Vision have led to their exploration in time-series representation learning. Self-supervised learning methods in this domain fall primarily into two categories: masking-based and dropping-based pre-training. Masking-based methods, such as PatchTST, randomly mask segments of time-series as zeros and then reconstruct them using learned representations \citep{TST, PatchTST}.  However, as shown in Figure~\ref{pretraining_tasks}.b, these methods suffer from data distribution shift, which introduces noise into the pre-training stage and degrades the quality of the learned representations. Unlike images and text data \citep{bert, vit}, distribution shift presents a larger challenge for time-series data, where the introduced zeros substantially deviate from the actual signals \citep{CRT}.  In contrast, Cross-Reconstruction Transformer (CRT) adopts a dropping-based method to circumvent distribution shift by discarding and reconstructing data segments \citep{CRT}. However, as shown in Figure~\ref{pretraining_tasks}.c, this method risks disrupting shapelets, which are discriminative subsequences of time-series data crucial for identifying the target class labels of the sequence \citep{shapelet, learn_shapelet}. As a potential solution, GPT adopts a next-token prediction task, which effectively preserves both the original distribution and time-series shapelets without altering the data (Figure~\ref{pretraining_tasks}.a). Despite GPT's efficacy in representation learning in diverse domains, its potential in time-series domain is still underexplored \citep{gpt1}.

\begin{figure}[t]
\begin{center}
\centerline{\includegraphics[width=0.7\columnwidth]{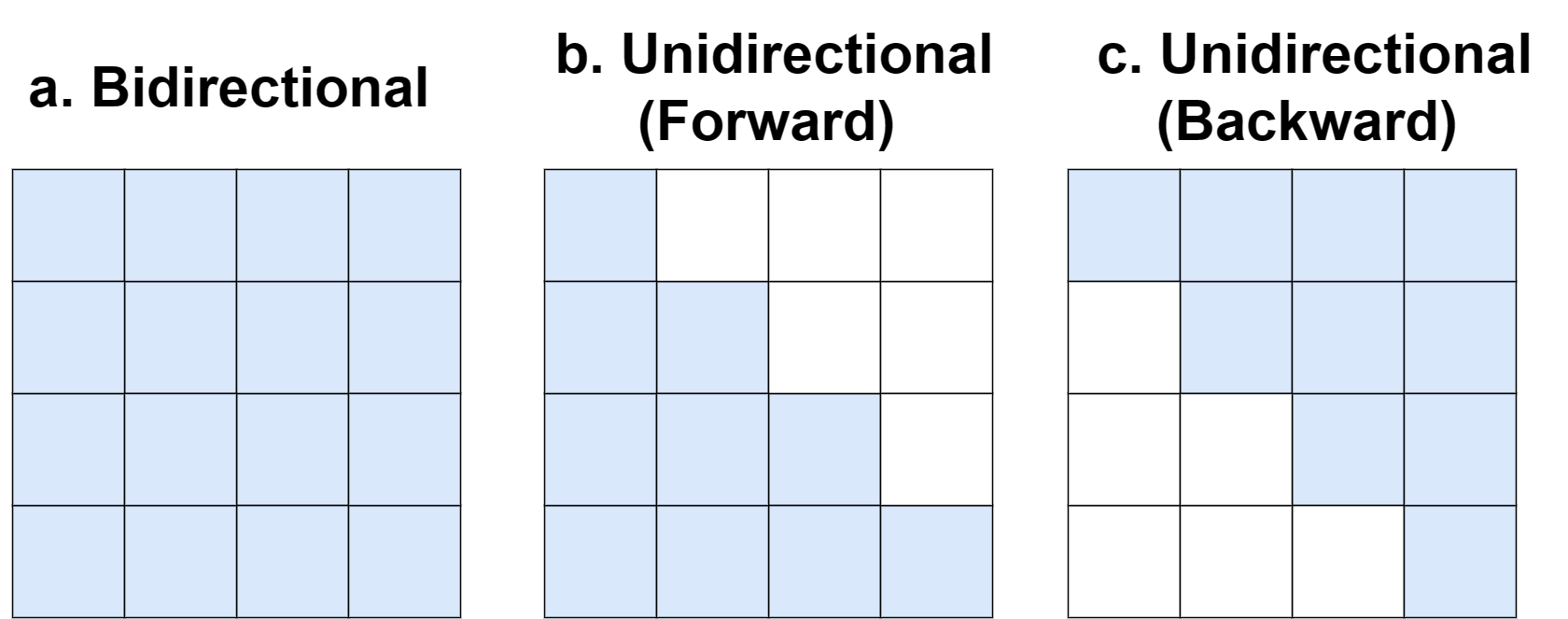}}
\caption{Bidirectional and unidirectional (forward and backward) attention matrices. Unlike bidirectional self-attention (\textbf{a}), \model alternates forward (\textbf{b}) and backward (\textbf{c}) attention across layers, achieving bidirectionality and full-rankedness. }
\label{fig:attention_matrix}
\end{center}
\vskip -0.4in
\end{figure}

While GPT is exceptionally effective for generative forecasting tasks using the unidirectional history contexts, its causal attention (Figure~\ref{fig:attention_matrix}.b) is less effective in discriminative tasks that require bidirectional contexts \citep{bert}. Using bidirectional self-attention mechanism (Figure~\ref{fig:attention_matrix}.a), PatchTST and CRT achieve superior performance in these discriminative tasks. While this self-attention mechanism captures bidirectional contexts, its attention matrix is low rank and thus faces a loss of expressiveness \citep{Low-Rank_Bottleneck}. These limitations underscore the necessity for a novel architecture that extracts bidirectional contexts with an expressive, full-rank attention matrix.

In this study, we aim to improve healthcare time-series representation learning via a novel pre-training strategy towards enhancing downstream seq2vec prediction tasks. To this end, we propose the \textbf{Bidirectional Timely Generative Pre-trained Transformer (\model)} that integrates bidirectionality into generative pre-training to improve time-series representation learning. It introduces a Next-Previous-Token Prediction pre-training task, preserving the original distribution and time-series shapelets without any data alterations (\cref{pretraining_tasks}.a).  \model alternates forward and backward attention across layers to pre-train deep bidirectional representations. Furthermore, both  forward and backward attention matrices in \model are full-rank and thus exhibit expressive representation power \citep{Low-Rank_Bottleneck}. The qualitative and quantitative results demonstrate the strengths of \model over existing methods (\cref{tab:model_comparison}), including data preservation, bidirectionality, and full-rankedness. The key contributions are threefold:
\begin{enumerate}
    \item  a novel Next-Previous-Token Prediction pre-training task preserves the original data distribution and time-series shapelets (\cref{pretraining_tasks});
    \item a novel \textbf{B}idirectionally \textbf{A}lternating \textbf{A}uto\textbf{R}egressive Modeling \textbf{(BAAR)} framework alternately models left-to-right and right-to-left information across layers, learning deep bidirectional contexts for discriminative tasks (\cref{architecture});
    \item the resulting forward and backward attention matrices, which are full-rank, exhibit expressive representation capabilities for pre-training (\cref{fig:attention_matrix}). 
\end{enumerate}

\subsection*{Generalizable insights about machine learning in the context of healthcare}

Healthcare time series provide rich opportunities for learning representations that enhance downstream medical applications, such as disease classification and physical activity prediction. Our study proposes \model to effectively pre-train temporal representations from biosignals and longitudinal clinical records and provides two key insights into modeling time-series health data:
    
\begin{enumerate}
    \item We have demonstrated the significance of modeling bidirectional contexts with full-rank attention matrices for predicting both continuously monitored physiological measurements and irregularly sampled diagnostic records. Our finding reveals that representation learning of complex temporal patterns can enhance downstream discriminative tasks. Future research could explore the generalizability of representation learning across different types of healthcare data and various downstream tasks.
    \item We have developed a flexible and efficient algorithm, \model, for representing deep bidirectional contexts in various healthcare time series data. This approach outperforms current state-of-the-art techniques in modeling and predicting physiological signals and longitudinal medical records. Future research can make use of this algorithm or design novel approaches based on our study to further enhance discriminative tasks in different healthcare datasets.
\end{enumerate}

\section{Related Work}

\subsection{Transformer architecture}
Transformer is an encoder-decoder model architecture stacked by $L$ layers of Transformer blocks \citep{transformer}. Each block consists of a self-attention layer followed by a feed-forward layer.  Given an input embedding $\mX \in \R^{N \times d}$, where $N$ is the number of tokens and $d$ is the hidden dimension, the output of a self-attention layer is computed as follows. 
\begin{equation}
\label{eq:attn_transformer}
    \text{Attention}(\mX) = \underbrace{\text{Softmax}\left( \frac{\mQ \mK^\top}{\sqrt{d}} \right)}_{\overleftrightarrow{\mA}} \mV
\end{equation}
where $\mQ,\mK, \mV = \mX \mW_Q, \mX \mW_K, \mX \mW_V \in \R^{N \times d}$ are Query, Key, and Value matrices, respectively. The bidirectional attention matrix $\overleftrightarrow{\mA}$ captures the context of each token based on all other tokens in the sequence.

Among various Transformer models, GPT and BERT stand out as decoder-only and encoder-only architectures, respectively \citep{gpt2, bert}. The unidirectional  GPT utilizes a causal self-attention mechanism:
\begin{equation}
\label{eq:attn_gpt}
    \text{CausalAttention}(\mX) = \underbrace{ \text{Softmax}\left( \frac{\mQ \mK^\top}{\sqrt{d}} \odot \mC \right)}_{\overrightarrow{\mA}}  \mV
\end{equation}
where both the mask matrix $\mC$ and forward attention matrix $\overrightarrow{\mA}$ are lower triangular (Figure~\ref{fig:attention_matrix}.b). GPT's Next-Token Prediction task effectively captures complex dependencies of natural language and thus excels in generative tasks. In contrast, BERT uses the self-attention (Figure~\ref{fig:attention_matrix}.a) and masking-based pre-training for deep bidirectional representations, which are crucial for discriminative tasks. Unlike BERT's bidirectional attention in each layer, \model alternates between forward and backward attention across layers (Figure~\ref{fig:attention_matrix}.b,c), learning deep bidirectional representations via a generative pre-training.

\subsection{Bidirectional recurrent representation learning}

Deep Bidirectional Long Short-Term Memory (BiLSTM) independently trains forward and backward LSTMs at each layer, followed by concatenating the hidden states from both directions post-layer. Embeddings from Language Models (ELMo) also trains forward and backward LSTMs independently, but concatenates their outputs only in the final layer \citep{ELMo}. Although ELMo effectively extracts pre-trained contextualized representations for downstream tasks, its bidirectional context is not deep due to the shallow concatenation of left and right contexts \citep{bert}. In contrast, we propose a BAAR framework that alternately models information flows from both directions across layers, allowing for \emph{deep} bidirectional representation learning.



\subsection{Recurrent Transformer RetNet}

Extrapolatable position (xPos) embedding encodes relative position information into Query $\tilde \mQ_n $ and Key $\tilde \mK_m$ based on the distances between tokens $n$ and $m$ \citep{xPos}:
\begin{align}
    & \tilde \mQ_n \tilde \mK_m^\top= \mX_n \mW_Q (\gamma e^{i\theta})^{n-m} (\mX_m \mW_K)^\top= \hat \mQ_n \gamma^{n-m}  \hat \mK_m^\top  \nonumber \\
    & \mathrm{with} \quad \hat \mQ_n =  \mX_n \mW_Q e^{i\theta n}, \hat \mK_m =  \mX_m \mW_K e^{-i \theta m}
\end{align}
where $\theta$ and $\gamma$ are rotation and exponential decay hyperparameters. The RetNet model introduces a Retention mechanism based on the xPos embedding \citep{retnet}:
\begin{equation}
\label{eq:Retention}
\overrightarrow{\text{Ret}}(\mX) = \underbrace{(\hat \mQ \hat \mK^\top \odot \overrightarrow{\mD})}_{\overrightarrow{\mR}} \mV, \: \overrightarrow{\mD}_{nm} = \begin{cases} \gamma^{n-m},   & n \geq m\\ 0, & n < m \end{cases}
\end{equation}
where $\overrightarrow{\mD} \in \R^{N \times N}$ is a lower triangular decay matrix. The Retention mechanism can be reformulated as an RNN for $n$-th timestep:
\begin{equation}
    \overrightarrow{\text{Ret}}(\mX_n) = \hat \mQ_n \mS_n, \:  \mS_n  = \hat \mK_n^\top \mV_n + \gamma \mS_{n-1} 
\end{equation}

TimelyGPT, built upon RetNet, performs long-term forecasting for biosignals and longitudinal clinical records \citep{timelygpt}. To adapt \emph{irregularity} in EHR data, TimelyGPT adapts the relative distance $n–m$ to account for the uneven gap between diagnoses, enabling it to compute time-decaying attention for these irregular gaps. Consequently, each element of the decay matrix $\overrightarrow{D}_{nm}$ in \cref{eq:Retention} reflects the relative distance between the irregularly sampled timesteps $n$ and $m$ in months. However, unidirectional TimelyGPT falls short in discriminative tasks compared to bidirectional CRT without the aid of convolution modules. To bridge this gap, we propose \model, which leverages a novel BAAR framework, to integrate bidirectionality into autoregressive time-series modeling.

\section{BiTimelyGPT Methodology}

\begin{figure}[t]
\centerline{\includegraphics[width=\textwidth]{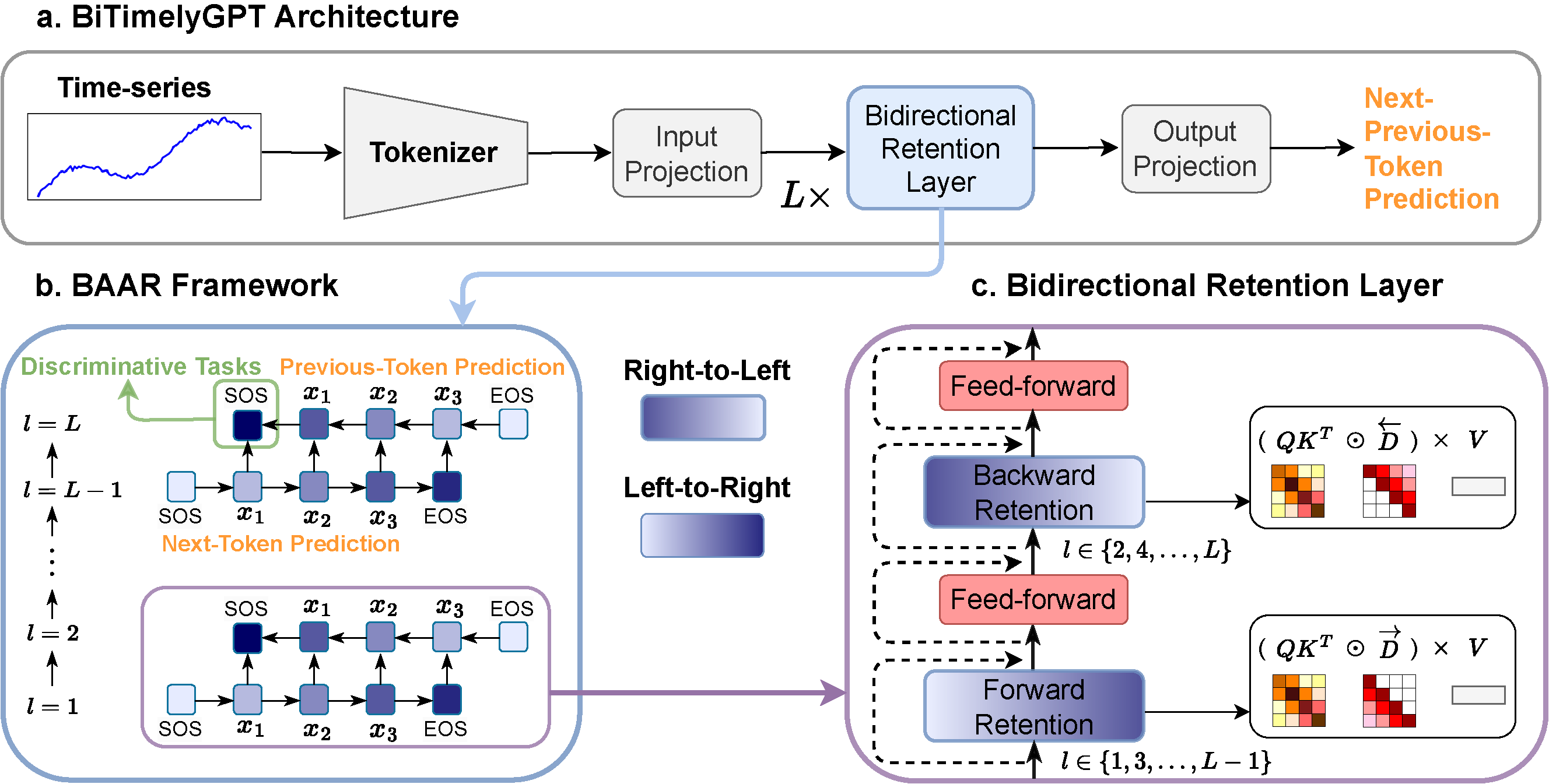}}
\caption{\model overview.  \textbf{a.} \textbf{\model architecture} is stacked with $L$ bidirectional Retention layers, with an overflow of pre-training in \cref{sec:overview}. \textbf{b.} \textbf{BAAR} framework (\cref{sec:BAAR}) alternately models left-to-right and right-to-left information across layers. It performs Next-Previous-Token Prediction on the final two layers and utilizes the \code{[SOS]} token in the final layer for discriminative tasks. \textbf{c.}  \textbf{Bidirectional Retention Layer} (\cref{sec:BiTimelyGPT}) alternates between forward and backward Retention across layers.
}
\label{architecture}
\vskip -0.1in
\end{figure}

An input sequence of time-series $x$ is denoted as $\{x_1, x_2, \ldots, x_T\}$ over $T$ timesteps, with  $x_t$ comprising $V$ features.  As depicted in Figure~\ref{architecture}.a, \model utilizes a convolution-subsampling tokenizer with two 1-D convolution layers, reducing the $T\times V$ down to an $N \times V$ sequence with $N < T$. These tokens are projected onto an input embedding $\mX \in \R^{N \times d}$, which is prepended and appended with \code{[SOS]} and \code{[EOS]} tokens, respectively. \cref{sec:overview} depicts the details. As stated above, the goal of time-series representation learning in this study is to pre-train temporal representations for fine-tuning seq2vec prediction tasks. Here, the output vector $\hat y$ can be either a continuous variable for regression or a categorical variable for classification. We describe the BAAR framework in \cref{sec:BAAR}, the \model pre-training in \cref{sec:BiTimelyGPT}, and the desired full-rank attention matrix as a result in \cref{sec:full_rank}. The notations of variables are defined in \cref{sec:notations}.

\subsection{Bidirectional Alternating AutoRegressive Modeling}
\label{sec:BAAR}

The BAAR framework presents a general approach  for incorporating bidirectionality into autoregressive models. It alternates between left-to-right and right-to-left information flows depending on the odd and even layer indices, respectively. In the odd-indexed layers $l \in \{1, 3, \ldots , L-1\}$, the forward model computes the probability of the sequence $x$ via a conditional probability of $x_t$ given the history context $(x_1, \ldots , x_{t-1})$ \citep{NIPS2000_728f206c}:
\begin{equation}
    \label{eq:forward_model}
    p(x_1, x_2, \ldots , x_T) = \prod_{t=1}^T p(x_t \mid x_1, \ldots , x_{t-1})
\end{equation}

This results in a history-dependent representation vector $\overrightarrow{h}_{t}^{(l)}$ given its history context $(x_1, \ldots , x_{t-1})$. In the even-indexed layers $l \in \{2, 4, \ldots , L\}$, the backward model runs over the sequence $x$ in reverse, predicting a timestep $x_t$ given the future context $(x_{t+1}, \ldots , x_T)$:
\begin{equation}
    \label{eq:backward_model}
    p(x_1, x_2, \ldots , x_T) = \prod_{t=1}^T p(x_t \mid x_{t+1}, \ldots , x_T)
\end{equation}

This yields a future-dependent representation vector $\overleftarrow{h}_{t}^{(l)}$ given its future context $(x_{t+1}, \ldots , x_T)$. The conditional probabilities in \cref{eq:forward_model} and \cref{eq:backward_model} can be modeled by forward and backward attention \citep{gpt1}. 

Figure~\ref{architecture}.b illustrates how the BAAR framework alternates between forward and backward layers to learn deep bidirectional representations. In the odd-indexed layer, the forward attention starts with a \code{[SOS]} token to predict the next tokens from left to right, where the \code{[EOS]} token summarizes information from all timesteps. In the even-indexed layer, the backward attention begins with the \code{[EOS]} token to predict the previous token in the right-to-left direction, intending to reconstruct the sequence in reverse. By alternating left-to-right and right-to-left information flows, the \code{[EOS]} and \code{[SOS]} tokens aggregate the sequence's features in the odd and even layers, respectively. In final layer (assuming an even number  $L$), the \code{[SOS]} token serves as a sequence representation for discriminative tasks.

\subsection{BiTimelyGPT architecture}
\label{sec:BiTimelyGPT}

As shown in Figure~\ref{architecture}.c, \model employs the BAAR framework to alternately model left-to-right and right-to-left information flows across layers. In the odd-indexed layers $l \in \{1, 3, \ldots , L-1\}$, \model utilizes the forward Retention to model left-to-right information with \cref{eq:Retention}, where the lower triangular decay matrix $\overrightarrow{\mD}$ maintains this direction. In the even-indexed layers $l = (2, 4, \ldots , L)$, \model utilizes the backward Retention to model right-to-left information:
\begin{equation}
\label{eq:backward_Retention}
\overleftarrow{\text{Ret}}(\mX)  = \underbrace{(\hat \mQ \hat \mK^\top \odot \overleftarrow{\mD}) }_{\overleftarrow{\mR}} \mV, \: \overleftarrow{\mD}_{nm} = 
    \begin{cases}
        \gamma^{m-n},   & n \leq m\\
        0, & n > m 
    \end{cases}
\end{equation}
where an upper triangular decay matrix $\overleftarrow{\mD}$ maintains right-to-left direction. Consequently, \model alternates between forward and backward Retention across layers, pre-training deep bidirectional representations.  



In \model, bidirectional generative pre-training aims to learn deep bidirectional representations from unlabeled data, preserving the original distribution and time-series shapelets without any data alterations. To maximize the joint probability of observing the entire sequence, \model predicts the next token in Layer $L-1$ and the previous token in Layer $L$. We define the objective of the Next-Previous-Token Prediction task as follows.
\begin{align}
    \sum_{t=1}^T & \log p(x_t^{(L-1)} \mid x_1^{(L-1)}, \ldots , x_{t-1}^{(L-1)})  + \log  p(x_t^{(L)} \mid x_{t+1}^{(L)}, \ldots , x_T^{(L)})
\end{align}

The output projection layer takes each token's representations $\overrightarrow{h}_{t}^{(L-1)}$ and $\overleftarrow{h}_{t}^{(L)}$ to predict the next and previous tokens, respectively. The pre-training loss is mean squared error (MSE) for continuous signals (e.g., biosignals) or cross-entropy for discrete signals (e.g., diagnosis codes). The \code{[SOS]} token in the final layer is used for fine-tuning as it aggregates a sequence representation for downstream tasks.

\begin{table}[t]
\centering
\caption{Comparison of models based on \model's three key advantages: data preservation, bidirectionality, and full-rankedness. }
\label{tab:model_comparison}
\footnotesize
\begin{tabular}{p{2.5cm}cccc} 
    \toprule
    \textbf{Model} & \textbf{Pre-training} & \textbf{Data Preservation} & \textbf{Bidirectionality} & \textbf{Full-rank} \\
    \midrule
    PatchTST & Masking-based & \xmark & \cmark & \xmark \\
    CRT & Dropping-based & \xmark & \cmark & \xmark \\
    TimelyGPT & Next-Token Pred & \cmark & \xmark & \cmark \\
    TimelyBERT & Masking-based & \xmark & \cmark & \xmark \\
    \textbf{BiTimelyGPT} &  Next-Previous-Token Pred
    & \textbf{\cmark} & \textbf{\cmark} & \textbf{\cmark} \\
    \bottomrule
\end{tabular}
\end{table}

\subsection{Full-rankedness in BiTimelyGPT}
\label{sec:full_rank}

The self-attention mechanism often suffers from the low-rank bottleneck, leading to a loss of expressiveness in the rank-deficient attention matrix $\overleftrightarrow{\mA} \in \mathbb{R}^{N\times N}$ \citep{attention_collapse, Low-Rank_Bottleneck}. For the single-head attention in \cref{eq:attn_transformer}, when the dimension $d$ is less than the sequence length $N$, we have $\mathrm{rank} (\mathbf{Q} \mathbf{K}^\top) \leq d$, resulting in a rank-deficient $\overleftrightarrow{\mA}$ without Softmax activation. This bottleneck persists even with non-linear Softmax operation \citep{wang2020linformer, mixhead}. This challenge is exacerbated in multi-head attention, where the dimension per head is much smaller than $N$ \citep{Low-Rank_Bottleneck}. In contrast, GPT's masked attention matrix  $\overrightarrow{\mA}$ in \cref{eq:attn_gpt}  is full-rank (i.e., $\mathrm{det}(\overrightarrow{\mA}) \neq 0$), forming a lower triangular attention matrix with strictly positive diagonal elements post-Softmax activation. Although this full-rank matrix $\overrightarrow{\mA}$ offers expressive representation power, it is limited to unidirectional contexts.

\model effectively overcomes the low-rank bottleneck by alternating between forward and backward Retention, resulting in triangular Retention matrices with non-zero diagonals. Specifically, the forward Retention matrix $\overrightarrow{\mR}$ in \cref{eq:Retention} becomes full-rank when its diagonal elements are non-zero, ensuring $\mathrm{det}(\overrightarrow{\mR}) \neq 0$. For a diagonal element $\overrightarrow{\mR}_{mm} = \mW_Q \mX_m (\mW_K \mX_m)^\top$, this condition is satisfied by initializing these embeddings to prevent orthogonality and avoid all-zero values across the $d$ dimensions. This reasoning also applies to the backward Retention matrix $\overleftarrow{\mR}$ in \cref{eq:backward_Retention}. Consequently, both forward and backward Retention matrices in \model are full-rank, enabling the learning of bidirectional contexts using more expressive attention matrices.

\subsection{Computational complexity}

\model with its efficient Retention mechanism achieves $O(N)$ complexity, in contrast to BERT and GPT, which incur $O(N^2)$ complexity \citep{linear_attention}. The self-attention mechanisms in \cref{eq:attn_transformer} and \cref{eq:attn_gpt} introduce a complexity of $O(N^2d)$. This quadratic computational bottleneck prevents standard Transformer models from modeling long sequences (i.e., $N >> d$). In contrast, \model achieves linear complexity by leveraging both forward and backward Retention mechanisms. For forward Retention, $\text{Ret}(X_n) =Q_n S_n, S_n=K_n^\top V_n + \gamma S_{n-1}$, where both $Q_n S_n$ and $K_n^\top V_n$ have $O(d^2)$ complexity. By recursively updating over $N$ timesteps, the total complexity becomes $O(N d^2)$. This is similarly applied in backward Retention. Therefore, \model ensures $O(2Nd^2)$ complexity, dramatically reducing computing resources when $N>>d$.

\section{Datasets}

\subsection{Biosignal datasets}

We used three publicly available, large-scale biosignal time-series datasets for pre-training: (1) the Sleep-EDF dataset with 7 types of biosignals across 1.2 billion timesteps \citep{sleep}; (2) the PTB-XL dataset with 12 types of electrocardiogram data totaling 109 million timesteps \citep{ptbxl}; (3) the PPG-Dalia dataset with 4 types of photoplethysmograph data from 16.6 million timesteps  \citep{DeepPPG}. Additionally, we employed five other datasets for downstream discriminative tasks, including Epilepsy \citep{Epilepsy}, EMG \citep{physiobank}, RR, HR, SpO2 \citep{Tan2020TSER}. The description and statistics of datasets are available in \cref{sec:data_des}.

\subsection{PopHR dataset}


Population Health Record (PopHR) was established to monitor population health in Montreal, Quebec, Canada \citep{pophr1, pophr2}. This healthcare administrative data includes clinical records of 1.3 million patients using International Classification of Diseases (ICD) codes. We extracted irregularly sampled time series from these clinical records. Specifically, we converted ICD-9 codes to phenotype codes (PheCodes) using the expert-defined PheWAS catalog \citep{phewas1, phewas2}. To identify relevant PheCodes for each target chronic disease, we calculated the prevalence ratio of each PheCode in the true patients compared to the general population. PheCodes with a prevalence ratio greater than 5 were included in our experiment. If no PheCode reached this threshold, we selected the top 5 PheCodes with the highest prevalence ratios.  We excluded patients with fewer than 10 PheCodes from the dataset. This resulted in an irregularly sampled time series dataset comprising 47,000 patients, 2.2 million records, and 47 unique PheCodes.


\section{Experiments}

\subsection{Qualitative analysis design on biosignals} 

To evaluate three key advantages of \model, we conducted qualitative analyses and interpreted visualization results on the large-scale Sleep-EDF dataset: (1) gradient-based saliency analysis for  data preservation during pre-training (\cref{sec:visual_pretraining});  (2) attention matrices for the effectiveness of bidirectionality (\cref{sec:visual_bidir}); (3) rank analysis of attention matrices for model expressiveness (\cref{sec:visual_rank}). To ensure consistency in comparison, we used the same patch-tokenizer with a non-overlapping patch size of 8, yielding $N=375$ tokens per 3000-timestep sequence \citep{PatchTST}.

\subsection{Quantitative experiment design on biosignals}

For the quantitative analysis in Section \ref{sec:cls_and_reg_exp}, we assessed \model's performance on downstream classification and regression tasks. For classification, we pre-trained on the Sleep-EDF and PTB-XL datasets separately. Subsequently, the PTMs were fine-tuned on the Sleep-EDF, Epilepsy, PTB-XL, and EMG datasets. For regression, we pre-trained on both the PTB-XL and PPGDalia datasets and then fine-tuned PTMs on the IEEEPPG, RR, HR, and SpO2 datasets. We divided each dataset into training (80\%), validation (10\%), and test (10\%) sets. When the same dataset was used for both pre-training and fine-tuning, 20\% of the training set was used for fine-tuning. All models adopted the preferred tokenization setup. We used accuracy and MAE as metrics for classification and regression, respectively.

\subsection{Experiment design on PopHR's irregularly sampled time series}
\label{sec:pophr_exp}

The PopHR database offers rule-based labels for three chronic phenotypes: congestive heart failure (CHF), chronic obstructive pulmonary disease (COPD), and diabetes. For downstream multi-label classification task in \cref{sec:pophr_exp}, we used cross entropy and area under precision recall curve (AUPRC) to evaluate pre-training and fine-tuning, respectively.

\subsection{Baselines} 

For the classification and regression tasks on biosignal datasets, we compared \model against a broad range of baselines, including transformer PTMs, consistency-based PTMs, and recurrent models. Among transformer models, we included TimelyGPT, TimelyBERT, CRT, PatchTST, AutoFormer \citep{autoformer}, FedFormer \citep{fedformer}, TST \citep{TST}, and TimesNet \citep{wu2023timesnet}. Consistency-based PTMs such as TS-TCC \citep{TS-TCC}, TS2Vec \citep{ts2vec}, and TF-C \citep{TF_C} were also evaluated.  We also assessed recurrent models like LSTM, BiLSTM, and ELMo. Additionally, to explore the efficacy of our proposed BAAR framework, we applied it to other autoregressive models, namely GPT-2 and LSTM. For PopHR's irregularly sampled time series, we compared \model with efficient PTMs from Section \ref{sec:cls_and_reg_exp}, including PatchTST, TST, CRT, TS-TCC, and TF-C. We also assessed algorithms designed for irregularly sampled time series, including mTAND \citep{shukla2021multitime}, GRU-D \citep{GRU_D}, SeFT \citep{SeFT}, and RAINDROP \citep{RAINDROP}.
We described the architectures and parameters for \model and baselines in \cref{sec: model_setup_details}. 



We also included a masking-based PTM TimelyBERT, which uses bidirectional Retention matrix $\overleftrightarrow{\mR}$ in each layer. This baseline modifies the decay matrix to $\overleftrightarrow{\mD} = \gamma^{|n-m|}$: 
\begin{equation}
\label{eq:bi_Retention}
\overleftrightarrow{\text{Ret}}(\mX) = \underbrace{(\hat \mQ \hat \mK^\top \odot \overleftrightarrow{\mD})}_{\overleftrightarrow{R}} \mV, \: \overleftrightarrow{\mD}_{nm} = \gamma^{|n-m|}
\end{equation}
However, the TimelyBERT baseline faces two limitations: (1) it relies on masking-based pre-training, leading to distribution shift; and (2) its bidirectional Retention $\overleftrightarrow{\mR}$ suffers from the low-rank bottleneck.

\subsection{Pre-training and fine-tuning}

\model performed pre-training with the Next-Previous-Token Prediction task, with fine-tuning on the \code{[SOS]} token in the final layer. For comparison, PatchTST adopted a masking-based method, masking 40\% of its patches to zero \citep{PatchTST}. CRT used a dropping-based method, discarding up to 70\% of patches \citep{CRT}. For transformers without established pre-training methods, we used a masking-based method by randomly masking 40\% of timesteps \citep{TST}. All transformers performed 20 epochs of pre-training, followed by 5 epochs of end-to-end fine-tuning.

\section{Results}

\subsection{Qualitative analysis of different pre-training strategies}
\label{sec:visual_pretraining}

We examined the impact of different pre-training methods on representational learning with respect to data preservation, focusing on \model, masking-based PatchTST, and dropping-based CRT. We assessed the saliency of the $N \times d$ output embedding matrix generated during pre-training by computing the absolute value of the corresponding gradients with respect to the prediction target \citep{saliency}. To identify discriminative patterns in time-series data \citep{shapelet, learn_shapelet}, we extracted shapelets from 10\% of the Sleep-EDF dataset using the pyts package and computed the average saliency values within these positions \citep{pyts}.

\begin{figure}[t]
\centerline{\includegraphics[width=\textwidth]{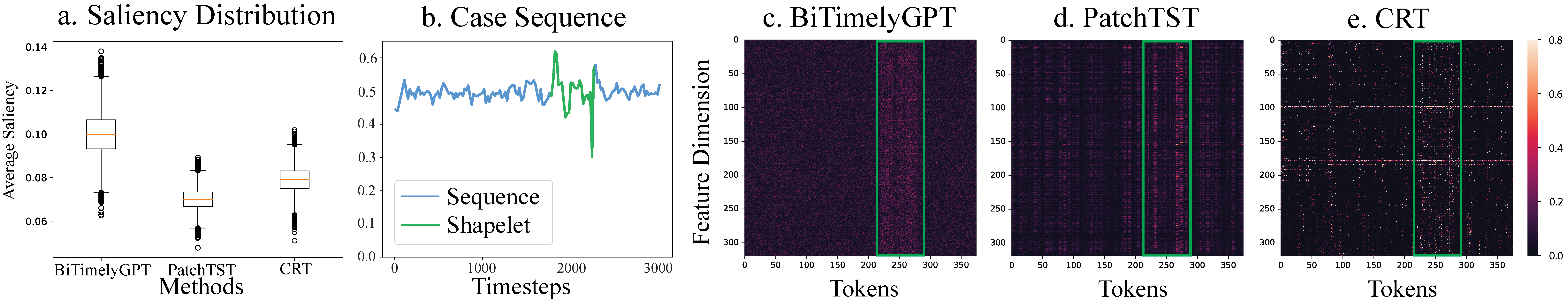}}
\caption{Visualization of different pre-training methods. \textbf{a.} Distribution of average saliency scores within the extracted shapelets of sequences.  \textbf{b.} A case example of a time-series and its shapelet. \textbf{c.} Generative pre-training of BiTimelyGPT. The green rectangle indicates the shapelet discriminative region. \textbf{d.}  Masking-based pre-training of PatchTST. \textbf{e.} Dropping-based pre-training of CRT. }
\label{fig:pretraining_saliency}
\end{figure}

As shown in Figure~\ref{fig:pretraining_saliency}.a, \model has higher saliency scores on the regions of data shapelets, suggesting its superior ability to uncover discriminative patterns compared to the baselines. Furthermore, we conducted a case study on a specific sequence with a distinct shapelet (Figure~\ref{fig:pretraining_saliency}.b). The saliency heatmap of the output embedding matrix (Figure~\ref{fig:pretraining_saliency}.c) reveals that \model effectively captured discriminative information surrounding the shapelet. In comparison, PatchTST (Figure~\ref{fig:pretraining_saliency}.d) and CRT (Figure~\ref{fig:pretraining_saliency}.e) display high saliency beyond the region of the shapelet driven by the minimization of the reconstruction loss, failing to concentrate on discriminative patterns. These observations underscored that \model's pre-training effectively discerns discriminative information compared to masking-based and dropping-based approaches. 
 
\subsection{Qualitative analysis of bidirectionality} \label{sec:visual_bidir}

\begin{figure}[t]
\centerline{\includegraphics[width=\textwidth]{
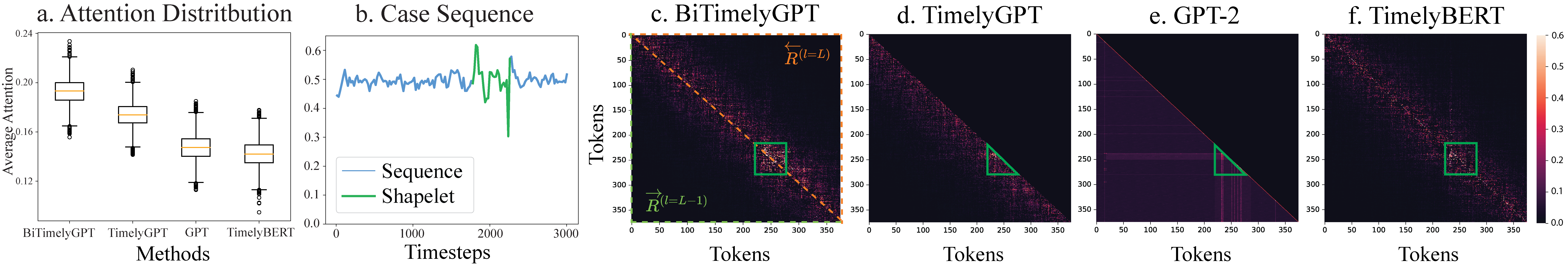}}
\caption{Visualization of bidirectionality advantage. \textbf{a.} Distribution of attention scores within the shapelet regions. \textbf{b.} A case example of time-series and its data shapelet. \textbf{c.} The combined Retention matrices from the last two layers of \model. The green rectangle indicates the shapelet discriminative region. \textbf{d.} TimelyGPT’s causal Retention matrix. \textbf{e.} GPT-2’s causal attention matrix. \textbf{f.} TimelyBERT's bidirectional Retention matrix.}
\label{bidirectionality}
\end{figure}

We investigated the effectiveness of bidirectionality in attention matrices in a downstream classification task. We compared the attention matrices of fine-tuned models of \model, TimelyGPT \citep{timelygpt}, GPT-2 \citep{gpt2}, and TimelyBERT. For \model, we combined its lower and upper triangular Retention matrices, $\overrightarrow{\mR}^{L-1}$ and $\overleftarrow{\mR}^{L}$, to demonstrate their bidirectional representations. The distribution of attention scores within the shapelet regions reveals  that \model achieves the highest scores on the discriminative regions (Figure~\ref{bidirectionality}.a). Given a case sequence, \model (Figure~\ref{bidirectionality}.c) exhibits localized and concentrated attention on the key discriminative region, efficiently integrating bidirectional contexts. In contrast, TimelyGPT (Figure~\ref{bidirectionality}.d) exhibits more limited focus on the discriminative segment. GPT-2 (Figure~\ref{bidirectionality}.e) has a broader but less precise range of attention around these key features. TimelyBERT (Figure~\ref{bidirectionality}.f), despite having decay mechanisms in both directions, struggles to concentrate attention on the discriminative region. These findings highlighted the effectiveness of \model in identifying discriminative patterns by modeling bidirectional contextualized representations.


\begin{figure}[b]
\begin{center}
\includegraphics[width=0.37\textwidth]{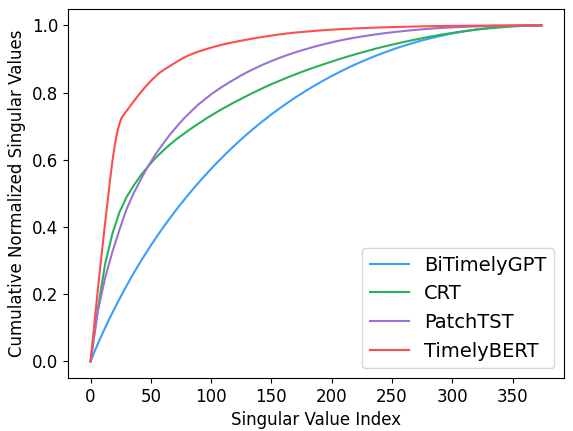}
\end{center}
\vspace*{-6mm}
\caption{Normalized cumulative singular values of the attention matrices. A more uniformly distributed curve of singular values indicates a higher rank attention matrix.}
\label{fig:svd}
\end{figure}

\subsection{Qualitative analysis of attention matrix rankedness}\label{sec:visual_rank}

We assessed the expressiveness of the attention matrices in various bidirectional baselines, including PatchTST, CRT, and TimelyBERT. We applied singular value decomposition to the attention matrices across different layers and heads, visualizing the normalized cumulative singular values (\cref{fig:svd}). \model exhibits a more uniform distribution of singular values, effectively overcoming the low-rank bottleneck. In contrast, the singular value curves for other bidirectional baselines skew toward the upper left corner, suggesting that the largest few singular values account for most information in the attention matrices. Notably, TimelyBERT has the most long-tailed spectrum distribution, due to its Retention matrix $\overleftrightarrow{\mR}=\hat \mQ \hat \mK^\top \odot \overleftrightarrow{\mD}$ lacking softmax's non-linearity. As a result, the more even distribution of singular values in \model indicates  more expressive representation power, breaking the low-rank bottleneck seen in other bidirectional baselines with skewed distributions.

\begin{table}[t]
    \caption{Comparison of \model against various baselines for downstream seq2vec tasks. Metrics are reported as average (standard error) from a bootstrap evaluation of variance. The highest values per task (column) are highlighted in bold.}
    \label{tab:summary}
    \scriptsize
    \begin{tabular}{p{1.9cm}|p{1.5cm}p{1.2cm}|p{1.2cm}p{1.2cm}|p{1.2cm}p{1.4cm}p{1.5cm}p{1.4cm}}
    \toprule
    Task & \multicolumn{4}{c|}{Classification (Accuracy \%) } & \multicolumn{4}{c}{Regression (MAE)} \\
    \midrule
    Pre-training & \multicolumn{2}{c|}{Sleep-EDF} & \multicolumn{2}{c|}{PTB-XL} & \multicolumn{4}{c}{PTB-XL \& PPGDalia} \\
    \midrule
    Fine-tuning &  Sleep-EDF & Epilepsy & PTB-XL & EMG & IEEEPPG & RR & HR & SpO2 \\
    \midrule
    BiTimelyGPT & \scalebox{0.9}{\textbf{90.4 (2.5)}} & \scalebox{0.9}{\textbf{93.0 (1.9)}} & 87.5 (2.7) & 95.4 (2.6) & \scalebox{0.9}{\textbf{26.1 (2.5)}} & 2.81 (0.12) & \scalebox{0.9}{\textbf{8.42 (0.24)}} & \scalebox{0.9}{\textbf{4.23 (0.11)}} \\
    TimelyGPT & 89.2 (3.9) & 92.8 (2.4) & 86.5 (3.2) & \scalebox{0.9}{\textbf{95.9 (3.3)}} & 26.2 (2.8) & \scalebox{0.9}{\textbf{2.78 (0.13)}} & 8.53 (0.19) & 4.26 (0.16) \\
    TimelyBERT & 82.2 (4.4) & 85.6 (3.2) & 81.0 (4.8) & 84.5 (3.3) & 31.0 (3.0) & 4.22 (0.22) & 12.86 (0.28) & 4.82 (0.19) \\
    CRT & 90.1 (3.2) & 91.1 (2.4) & \scalebox{0.9}{\textbf{87.8 (3.2)}} & 94.6 (3.1) & 26.5 (2.6) & 2.96 (0.17) & 9.02 (0.22) & 4.48 (0.13) \\
    PatchTST & 89.6 (3.4) & 91.3 (2.5) & 83.4 (3.2) & 95.2 (3.0) & 26.1 (2.7) & 2.89 (0.12) & 9.46 (0.24) & 4.45 (0.13) \\
    AutoFormer & 78.9 (3.6) & 84.2 (2.5) & 78.5 (3.5) & 88.6 (3.7) & 32.2 (2.9) & 4.13 (0.15) & 13.29 (0.28) & 4.95 (0.13) \\
    FedFormer & 76.4 (3.6) & 81.7 (2.7) & 75.5 (3.5) & 85.2 (2.7) & 31.1 (2.5) & 4.36 (0.21) & 13.82 (0.32) & 4.75 (0.15) \\
    TimesNet & 83.6 (3.5) & 86.0 (2.7) & 79.4 (3.2) & 89.3 (3.2) & 29.9 (2.7) & 4.19 (0.18) & 13.65 (0.27) & 4.83 (0.16) \\
    TST & 88.8 (3.5) & 88.0 (2.5) & 81.9 (3.2) & 94.2 (2.9) & 26.8 (2.7) & 3.47 (0.13) & 12.63 (0.27) & 4.95 (0.15) \\
    GPT & 84.3 (3.1) & 88.7 (2.6) & 83.1 (3.1) & 92.6 (2.7) & 28.6 (2.8) & 3.71 (0.12) & 11.09 (0.30) & 4.64 (0.18) \\
    GPT-BAAR & 86.2 (3.5) & 89.7 (2.8) & 85.1 (3.1) & 93.7 (3.3) & 28.2 (2.8) & 3.60 (0.19) & 10.62 (0.27) & 4.66 (0.13) \\
    \hline
    TS2Vec & 86.2 (3.8) & 88.3 (3.1) & 82.7 (4.1) & 93.8 (3.3) & 27.9 (2.9) & 3.53 (0.24) & 11.56 (0.32) & 4.60 (0.19) \\
    TS-TCC & 86.1 (3.9) & 89.7 (2.9) & 84.7 (3.9) & 93.3 (3.8) & 29.3 (3.2) & 4.09 (0.23) & 13.64 (0.35) & 4.86 (0.19) \\
    TF-C & 86.6 (3.9) & 87.5 (2.8) & 82.7 (3.7) & 93.8 (3.5) & 28.5 (3.9) & 4.38 (0.19) & 14.15 (0.29) & 4.87 (0.20) \\
    \hline
    LSTM & 80.2 (3.9) & 76.5 (3.2) & 78.8 (3.5) & 87.0 (2.9) & 30.2 (3.2) & 4.95 (0.22) & 14.37 (0.41) & 5.05 (0.23) \\
    LSTM-BAAR & 82.2 (3.5) & 81.1 (2.9) & 82.3 (3.1) & 88.0 (3.0) & 28.1 (2.8) & 4.53 (0.14) & 11.78 (0.28) & 4.82 (0.19) \\
    BiLSTM & 81.6 (3.5) & 78.3 (3.3) & 80.7 (3.5) & 86.8 (2.8) & 29.4 (2.8) & 4.71 (0.17) & 13.50 (0.29) & 4.97 (0.21) \\
    ELMo & 82.2 (3.5) & 81.6 (3.0) & 81.0 (3.1) & 88.5 (2.9) & 28.2 (3.0) & 4.66 (0.15) & 12.42 (0.24) & 4.76 (0.15) \\
    \bottomrule
    \end{tabular}
\end{table}

\subsection{Quantitative results on discriminative fine-tuning tasks}
\label{sec:cls_and_reg_exp}

As shown in \cref{tab:summary}, \model demonstrates the best performance in classification on the Sleep-EDF and Epilepsy datasets as well as regression on the IEEEPPG, HR and SpO2 datasets. These results highlighted the effectiveness of \model's bidirectionality, surpassing unidirectional TimelyGPT. Additionally, data preservation during pre-training and full-rank Retention matrices contribute considerably to \model's superior performance over other bidirectional transformers. CRT, employing a dropping-based pre-training, performs consistently well across datasets and achieves top results in the PTB-XL dataset. In contrast, masking-based PTMs typically lag behind due to the distribution shift. Among these masking-based PTMs, Autoformer, Fedformer, and TimesNet, which rely heavily on time decomposition and frequency-domain information, are especially affected by distribution shift. 
This suggested the advantage of dropping-based pre-training regime in avoiding distribution shift over the masking-based pre-training methods.
Furthermore, both consistency-based PTMs and recurrent models show inferior performance in these tasks. The applications of the BAAR framework to GPT and LSTM result in improved performance, indicating its broad adaptability and effectiveness in enhancing various autoregressive models.

\subsection{Classification of chronic disease from PopHR's longitudinal records}
\label{sec:cls_pophr}

\model achieved superior performance over other time-series PTMs for classification of CHF, COPD, and Diabetes phenotypes (Table~\ref{pophr_cls}), which is attributable to its ability of utilizing bidirectional contexts. This proficiency in handling bidirectional information from irregularly sampled time series highlighted \model's potential in healthcare applications, where patients' longitudinal records are often sporadic due to the irregular nature of patient visits. Notably, our \model also surpasses other methods designed for irregularly sampled time series, leading with an average AUPRC of 60.5\% and outperforming the next best mTAND (57.2\%). Such results  underscored the \model's generalizability, affirming its capability to model both continuous and irregularly sampled time series.

\begin{table*}[t]
\caption{Comparison of TimelyGPT and 9 baselines for multi-label classification (AUPRC \%) on the three chronic diseases. Metrics are reported as average (standard error) from bootstrap evaluation of variance. Best values are highlighted in bold.}
\vspace*{-5mm}
\small 
\label{pophr_cls}
\begin{center}
\footnotesize
\begin{tabular}{l|c|c|c}
\hline
\textbf{Model} & \textbf{CHF} & \textbf{COPD} & \textbf{Diabetes} \\
\hline 
\model & \bf 65.2 (2.2)  & \bf 54.8 (1.6)  &  \bf 61.4 (1.4)  \\
TimelyGPT & 63.7 (3.3)  & 54.1 (2.8)  &  56.3 (2.4)  \\
PatchTST & 58.5 (3.5)  & 48.9 (2.9)  & 51.2 (2.2)  \\
TST & 59.3 (3.2)  & 49.7 (2.3) & 52.0 (2.7)  \\
CRT & 54.2 (3.3)  & 44.6 (3.2) & 46.7 (2.6)  \\
TS-TCC & 52.6 (4.2) & 43.0 (3.9) & 45.3 (3.7) \\
TS-T & 50.5 (4.8) & 40.9 (4.2) & 43.1 (4.0)  \\
mTAND & 61.6 (1.9)  & 54.7 (1.6) & 55.3 (1.4) \\
GRU-D & 59.9 (2.6) & 53.3 (2.1) & 49.5 (2.0) \\
SeFT & 54.8 (2.4)  & 45.2 (2.0) & 47.5 (1.9)\\
RAINDROP & 58.2 (3.1) & 51.6 (2.5) & 52.8 (2.6) \\
\hline
\end{tabular}
\end{center}
\vskip -0.2in
\end{table*}

\subsection{Ablation study}

To evaluate different pre-training strategies while fixing the same neural architecture, we conducted ablation studies using the Sleep-EDF and PopHR datasets for classify sleep stages and CHF, respectively. We replaced the BAAR framework with two alternative pre-training strategies, namely (1) next-token prediction; (2) masking-based pre-training. We excluded the dropping-based method, which is tailored for encoder-decoder architectures. As shown in \cref{tab:ablation_cls}, the next-previous token prediction strategy achieves the highest performance, with 90.4\% for Sleep-EDF and 65\% for CHF. Replacing the BAAR framework with next-token prediction and masking-based pre-training resulted in a decline in performance for both datasets. Notably, the masking-based TimelyBERT underperforms compared to \model even when our model is trained without pre-training, showing decreases of 4.1\% for Sleep-EDF and 6.5\% for CHF. This suggested that the Retention mechanism, which introduces decay in two directions, has less expressive representation power as discussed in \cref{sec:visual_rank}. These findings highlighted the contribution of the BAAR framework and next-previous token Prediction in enhancing performance on discriminative tasks.


\begin{table}[h]
\centering
\caption{Ablation results of \model by assessing different pre-training strategies in terms of downstream classification tasks on Sleep-EDF and PopHR datasets.}
\label{tab:ablation_cls}
\footnotesize
\begin{tabular}{lcc}
\toprule
\textbf{Pre-training Strategies} & \textbf{Sleep-EDF}  & \textbf{CHF (PopHR)} \\
\midrule
\model with next-previous token prediction (Ours) & 90.4 & 65.2  \\
Replacing BAAR with next token prediction & 87.7 &  63.2 \\
Replacing BAAR with masking-based pre-training (i.e., TimelyBERT) & 82.2 & 57.8 \\ 
\model without next-previous token prediction & 86.3 & 64.3 \\
\bottomrule
\end{tabular}
\end{table}

\subsection{Probing study}
We also explored the most effective choice of sequence features for downstream classification tasks using Sleep-EDF and PopHR datasets (\cref{tab:ablation_token}). Focusing on the last layer ($l=L$) and the last two layers ($l=\{L-1, L\}$), we examined four token combinations: \code{[SOS]} token alone, \code{[EOS]} token alone, both \code{[SOS]} and \code{[EOS]} tokens, and average pooling of all tokens. We observed that the \code{[SOS]} token from the last layer achieves the best classification results, which can be attributed to its aggregation of sequence information by backward attention.

\begin{table}[t]
\centering
\caption{Probing study on sentence representation for downstream classification tasks using Sleep-EDF and PopHR datasets, showing the best results  with \code{[SOS]} in the final layer. }
\label{tab:ablation_token}
\footnotesize
\begin{tabular}{lcc}
\toprule
\textbf{Sequence Represntation} & \textbf{Sleep-EDF} & \textbf{CHF (PopHR)} \\
\midrule
\textbf{\code{[SOS]} } ($l=L$) & 90.4 & 65.2 \\
\text{\code{[EOS]} } ($l=L$) & 89.7 &  64.5 \\
\text{\code{[SOS]}+ \code{[EOS]} } ($l=L$) & 90.1 & 64.9  \\
\text{Average Pooling ($l=L$)} & 90.2 & 64.7\\
\hline
\text{\code{[SOS]} } ($l=\{L-1, L\}$) & 88.6 & 64.4  \\
\text{\code{[EOS]} } ($l=L$) & 89.5  & 64.6 \\
\text{\code{[SOS]} + \code{[EOS]} } ($l=\{L-1, L\}$) & 89.2 & 64.6 \\
\text{Average Pooling  ($l=\{L-1, L\}$)}  &  89.6 & 64.4 \\
\bottomrule
\end{tabular}
\end{table}

\section{Discussion} 

We present a \model model which leverages the recurrent attention and BAAR framework, for time-series representation learning and downstream discriminative tasks. Extensive visualization analyses underscore \model's three contributions: (1) the Next-Previous-Token Prediction pre-training effectively preserves the original distribution and data shapelets; (2) the BAAR framework alternates forward and backward Retention across layers, learning deep bidirectional representations; (3) the full-rank forward and backward Retention matrices enhance representation expressiveness. Empirical results demonstrate \model's superiority in classification and regression on eight datasets. 

\paragraph{Limitations}

This study concentrates on in-domain healthcare time-series data sharing similar biological patterns; however, its capacity to transfer learned representations to out-of-distribution biosignals remains untested. Future research should explore these out-of-distribution biosignals and data modalities to fully examine the model's transferability  across a wide range of healthcare data. Furthermore, while the BAAR framework's effectiveness has been demonstrated with the Retention mechanism, its potential benefits for a variety of autoregressive models need further exploration.

\section{Acknowledgements}
Y.L. is supported by Canada Research Chair (Tier 2) in Machine Learning for Genomics and Healthcare (CRC-2021-00547), NSERC Discovery Grant (RGPIN-2016-05174), NOVA–FRQNT-NSERC (2023-NOVA-328677). Z.S. is supported by FRQ-NT Doctoral Training Scholarships.

\clearpage


\bibliography{sample}

\newpage
\appendix

\section{Additional Information for \model}

\subsection{Denotations of Variables}
\label{sec:notations}

\begin{table}[h] 
\caption{Notations in \model}
\label{table: var_description}
\vskip 0.15in
\begin{center}
\scriptsize
\begin{tabular}{ l | l || l | l }
\hline
\textbf{\normalsize Notations}  & \textbf{\normalsize Descriptions} & \textbf{\normalsize Notations}  & \textbf{\normalsize Descriptions} \\
\hline 
$T$ & Number of timesteps & $N$ & Number of tokens \\ 
$V$ & Number of variates  & $L$ & Number of layers \\
$d$ & Hidden dimension & $h$ & Number of heads \\
$x \in \R^{T \times V} $ & A time-series sequence & $\mX \in \R^{N \times d} $ & A sequence of tokens \\
\code{[SOS]} & Start token of a sequence & \code{[EOS]} & End token of a sequence \\
$\mQ, \mK, \mV \in \R^{N \times d} $ & Query, key, value matrices & $\mW_Q, \mW_K, \mW_V \in \R^{d \times d} $ & Projection matrices for $\mQ, \mK, \mV$  \\ 
$\overleftrightarrow{\mA} \in \R^{N \times N} $ & Bidirectional attention matrix & $\overrightarrow{\mA} \in \R^{N \times N} $ & Forward attention matrix \\ 
$\mC\in \R^{N \times N} $ & Lower triangular mask matrix &  $\mS \in \mR^{d \times d}$ & State variable for retention \\ 
$\tilde \mQ, \tilde  \mK, \in \R^{N \times d} $ & $\mQ, \mK, \mV$  with xPos embedding & $\hat \mQ, \hat  \mK, \in \R^{N \times d} $ & $\mQ, \mK, \mV$ with RoPE embedding \\
$\theta $ & Rotary angle hyperparameter & $\gamma $ & Exponential decay hyperparameter \\
$\overrightarrow{\mR} \in \R^{N \times N} $ & Forward retention matrix & $\overrightarrow{\mD} \in \R^{N \times N} $ & Lower triangular decay matrix \\
$\overleftarrow{\mR} \in \R^{N \times N} $ & Backward retention matrix & $\overleftarrow{\mD} \in \R^{N \times N} $ & Upper triangular decay matrix\\ 
$\overleftrightarrow{\mR} \in \R^{N \times N} $ & Bidirectional retention matrix & $\overleftrightarrow{\mD} \in \R^{N \times N} $ & Bidirectional decay matrix  \\ 
\toprule
\end{tabular}
\end{center}
\vskip -0.1in
\end{table}

\subsection{BiTimelyGPT Pre-training overview}
\label{sec:overview}

\begin{figure}[h]
\begin{center}
\includegraphics[width=\textwidth]
{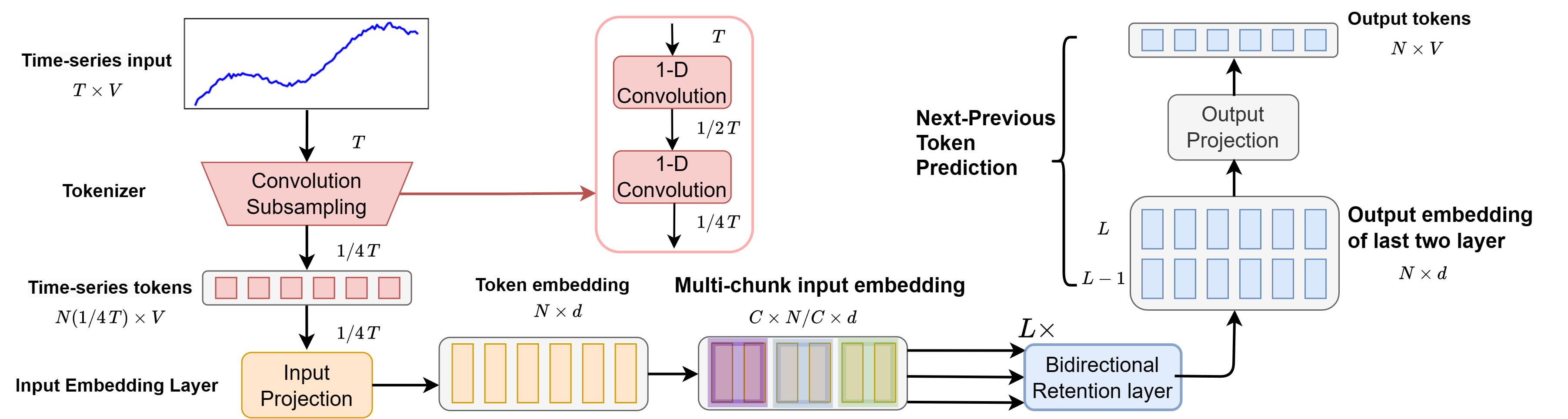}
\end{center}
\caption{Schematic of the \model Pre-Training Process}
\label{bitimelygpt_flow}
\end{figure}

\cref{bitimelygpt_flow} depicts the process of input processing followed by Next-Previous-Token Prediction pre-training. The input of time-series sequence with $T$ timesteps and $V$ features, $x\in \R^{T\times V}$ is tokenized via a convolution-subsampling module. The convolution-subsampling tokenizer comprises of two 1-D convolution layers with a kernel size of 3 and stride of 2 \citep{timelygpt}. The resulting sequence of token has dimension $N \times V$, reducing the sequence length by 1/4, i.e, $N = T/4$. These tokens are projected onto the input embedding $\mX \in \R^{N \times d}$ by an input projection layer. By adding \code{[SOS]} and \code{[EOS]} tokens, the sequence dimension becomes $(N+2)\times d$. Given the $L$ bidirectional generative layers, \model alternates between forward and backward Retention layers to train bidirectional contextualized representations. Moreover, the Retention mechanism provides an efficient chunk-wise forward pass that segments an input sequence into multiple chunks \citep{retnet}. Given a chunk size of $C$, the $N \times d$ input embedding is reshaped into a $C \times N/C \times d$ tensor. The output projection layer takes the output embedding with the shape $N \times d$ in the last two layers, which are used to predict the original sequence of tokens with the shape $N \times V$ for Next-Previous-Token Prediction task.

\section{Additional Information for Experiments}




\subsection{Dataset Description}
\label{sec:data_des}



\textbf{Sleep-EDF dataset for Visualization. }
Sleep-EDF dataset proposed by \citep{sleep} from PhysioBank \citep{physiobank} contains sleep cassette data obtained from 153 subjects. The collected whole-night polysmnographic (PSG) sleep recordings encompass 7 features: Electroencephalogram (EEG) (from Fpz-Cz and Pz-Oz electrode locations), electrooculogram (EOG) (horizontal), submental chin electromyogram (EMG), and an event marker. EEG and EOG signals were sampled at 100 Hz, while the EMG and event marker were sampled at 1 Hz. The sleep patterns correspond to the PSGs consist of five sleep stages: Wake (W), Non-rapid eye movement (N1, N2, N3) and Rapid Eye Movement (REM). This dataset contains a total of 1.2B timesteps, segmented into 400.7K sequences of 3,000 timesteps each. Its pre-training was utilized for visualization analyses detailed in \cref{sec:visual_pretraining}, \cref{sec:visual_bidir}, and \cref{sec:visual_rank}.

\noindent \textbf{Sleep-EDF (Sleep-stage only) dataset for Classification. }
We leveraged only the sleep stage of the Sleep-EDF dataset for the classification task following existing studies \citep{TS-TCC, TF_C}. In line with conventional practices, we selected the single EEG channel that captures signals from the Fpz-Cz electrode location. Since the Sleep-EDF dataset labels five sleep stages (i.e., W, REM, N1, N2 and N3), we only focused on EEG signals associated with these sleep stages, resulting in a total of 586.4 million timesteps. This dataset was segmented into 195.5K sequences, each with 3,000 timesteps.

\noindent \textbf{Epilepsy Seizure Classification. } Epileptic Seizure Recognition dataset \citep{Epilepsy} comprises EEG measurements from 500 subjects. This dataset captures brain electrical activity from different regions and states, and are divided into segments of 23.6 seconds. The original dataset consists of five classes of EEG measuring eyes open, eyes closed, healthy brain region, tumor region, and seizure. The first four classes were  merged into a single class as these classes are unrelated to epileptic seizure, enabling a binary classification of epileptic seizures. The dataset for classification consists of 2M timesteps, divided into 11.5K sequences, each containing 178 timesteps.

\noindent \textbf{PTB-XL Classification. } Physikalisch Technische Bundesanstalt large scale cardiology database (PTB-XL) \citep{ptbxl} from PhysioBank \citep{physiobank} contains 21,837 clinical 12-lead ECG signals (male: 11,379 and female: 10,458) with a duration of 10 seconds each, sampled with a rate of 500 Hz. These signals are categorized based on a set of twelve leads (I, II, III, AVL, AVR, AVF, V1, V2, V3, V4, V5, V6), where the reference electrodes on the right arm are provided for each signal, resulting in twelve labels for classification. This dataset for classification comprises 21,837 samples, each spanning 5,000 timesteps, amounting to a total of 109.2M timesteps.

\noindent \textbf{EMG Classification. } EMG dataset from PhysioBank \citep{physiobank} captures the electrical activity resulting from neural stimulation of muscles. This dataset provides insights about muscle functionality and the corresponding nerves responsible for their control. This dataset contains single-channel EMG signals sampled with a rate of 4 KHz. EMG recordings were obtained from the tibialis anterior muscle of volunteers exhibiting different degrees of muscular and neural disorders, resulting in three classification labels. This dataset for classification comprises 306.0K timesteps, segmented into 204 sequences, each consisting of 1,500 timesteps.

\noindent \textbf{PPG-Dalia Regression. } PPG-Dalia dataset \citep{DeepPPG} is available from Monash University, UEA\&UCR Time Series Regression Archive \citep{Tan2020TSER}. This dataset captures photoplethysmograph (PPG) data for motion compensation and heart rate estimation. Data was collected from 15 subjects engaging in daily life activities  using both wrist-worn (Empatica E4) and chest-worn (RespiBAN Professional) devices. All signals were sampled at 700 Hz. The ECG recording serve as a ground truth for heart rate. The PPG and 3D-accelerometer data are used to estimate heart rate, while accounting for motion artefacts. This dataset with a total of 16.6M timesteps was divided into 64,697 sequences, with varied lengths of 256 or 512 timesteps in different dimensions. The pre-training on this PPG-Dalia  as well as the PTB-XL dataset was utilized for fine-tuning regression tasks. 

\begin{table}[t]
\caption{The statistics of datasets. Note that the sequence length of the PPG-Dalia dataset varies across different dimensions. 
}
\label{transformer_setup_transposed}
\begin{center}
\footnotesize
\begin{tabular}{lcccccc}
\hline
{\bf Dataset} & {\bf Task} & {\bf Features} & {\bf Timesteps} & {\bf Sequences} & {\bf Length} & {\bf Classes} \\
\hline 
Sleep-EDF & Visualization & 7 & 1.2B & 400.7K & 3,000  & None \\
Sleep-EDF (sleep)  & Classification & 1 & 586.4M & 195.5K & 3,000 & 5 \\
Epilepsy & Classification & 1 & 2.0M & 11.5K & 178 & 2 \\
PTB-XL & Classification & 12 & 109.2M & 21,837 & 5,000 & 5 \\
EMG & Classification & 1 & 306.0K & 204 & 1,500 & 3 \\
PPG-Dalia & Regression & 4 & 16.6M & 64,697 & 256 \& 512$^*$ & None  \\
IEEEPPG & Regression & 5 & 3.1M  &  3,096 & 1,000 & None  \\
RR & Regression & 2 & 31.5M & 7,870 & 4,000 & None \\
HR & Regression & 2 & 31.8M & 7,949 & 4,000 & None\\
SpO2 & Regression & 2 & 31.8M & 7,949 & 4,000 & None\\
\hline
\end{tabular}
\end{center}
\end{table}

\noindent \textbf{IEEEPPG Regression. } IEEEPPG dataset \citep{IEEEPPG} is sourced from Monash University, UEA\&UCR Time Series Regression Archive \citep{Tan2020TSER}. This dataset aims to estimate  heart rate by utilizing PPG and ECG signals. The dataset includes two-channel PPG signals, three-axis acceleration signals, and one-channel ECG signals. These signals were recorded simultaneously and sampled at a rate of 125 Hz. This dataset for regression contains 3.1M timesteps, segmented into 3.096K sequences of 1,000 timesteps each.

\noindent \textbf{RR, HR and SpO2 Regression. } BIDMC Respiratory Rate (RR), heart rate (HR) and blood oxygen saturation level (SpO2) datasets \citep{RR} are available at Monash University, UEA\&UCR Time Series Regression Archive \citep{Tan2020TSER}. These datasets were obtained from the Physionet's BIDMC PPG and Respiration dataset \citep{physiobank}. They contain PPG and ECG signals sampled at a rate of 125 Hz, which are designed for estimating  RR, HR, and SpO2. The three datasets for regression were segmented into sequences of 4,000 timesteps each, yielding: (1) 7,870 sequences in the RR dataset, totaling 31.5M timesteps;  (2) 7,949 sequences in the HR dataset, encompassing 31.8M timesteps;  (1) 7,949 sequences in the SpO2 dataset, also amounting to 31.8M timesteps.

\noindent \textbf{PopHR database. } The database hosts a massive amount of longitudinal heterogeneous claim data from the provincial government health insurer in Quebec, Canada (Régie de l’assurance maladie du Québec, RAMQ) on health service use \citep{pophr1, pophr2}. In total, there are approximately 1.3 million participants in the PopHR database, which represents a randomly sampled 25\% of the population in the metropolitan area of Montreal between 1998 and 2014. Cohort membership is maintained dynamically by removing deceased residents and actively enrolling newborns and immigrants.

\subsection{Experiment Details for \model and Baselines} \label{sec: model_setup_details}

Given the demonstrated scaling law in the time-series datasets \citep{timelygpt}, we tailored the hyperparameters and model parameters corresponding to the size of pre-training datasets: (1) 18 million model parameters for the SleepEDF dataset; (2) 7.5 million model parameters for the PTB-XL dataset; (3) 7.5 million model parameters for the regression for the PTB-XL and PPGDalia datasets. For a fair comparison, we set the same model parameters as well as architectures for \model, transformer baselines, and recurrent models, as shown in \cref{BiTimelyGPT_baseline_setup}.

\begin{table*}[h]
\caption{Configurations of \model , transformer baselines, and recurrent models across different experiments and datasets}
\label{BiTimelyGPT_baseline_setup}
\begin{center}
\footnotesize
\begin{tabular}{lcccc}
\toprule
Pre-training Datasets & \bf Sleep-EDF & \bf PTB-XL & \bf PTB-XL\&PPGDalia & \bf PopHR \\
\midrule
Downstream Task & Classification & Classification & Regression & Classification \\
Data Size (Timesteps) & 1.2B & 109.2M & 109.2M \& 16.6M & 2.2M \\
Model Parameters & 18M & 7.5M & 7.5M & 3M \\
\hline
\multicolumn{5}{l}{\bf BiTimelyGPT} \\
\hline
Decoder Layers & 12 & 8 & 8 & 4 \\
Heads & 8 & 8 & 8 & 4 \\
Dim ($\mQ$, $\mK$, $\mV$, FF) & 320,320,640,640 & 240,240,480,480 & 240,240,480,480 & 144,144,288,288 \\
\hline
\multicolumn{5}{l}{\bf Transformer Baselines including Encoder-decoder, Encoder-only, and Decoder-only} \\
\hline
Enc-Dec Layers & 6 \& 6 & 4 \& 4  & 4 \& 4 & 4 \& 4 \\
Encoder Layers & 12 & 8 & 8 & 8 \\
Decoder Layers & 12 & 8 & 8 & 8 \\
Heads & 8 & 8 & 8 & 4 \\
Dim ($\mQ$, $\mK$, $\mV$, FF)  & 384,384,384,1536 & 288,288,288,1152 & 288,288,288,1152 & 144,144,144,576 \\
\hline
\multicolumn{5}{l}{\bf Recurrent Models} \\
\hline
Layers & 12 & 8  & 8 & 8 \\
Dim & 384 & 288 & 288 & 144 \\
\bottomrule
\end{tabular}
\end{center}
\end{table*}

\end{document}

%% file: math_commands.tex
\usepackage{amsmath,amsfonts,bm}

\def\eqref#1{equation~\ref{#1}}

\def\1{\bm{1}}

\def\mA{{\bm{A}}}

\def\mC{{\bm{C}}}
\def\mD{{\bm{D}}}

\def\mK{{\bm{K}}}

\def\mQ{{\bm{Q}}}
\def\mR{{\bm{R}}}
\def\mS{{\bm{S}}}

\def\mV{{\bm{V}}}
\def\mW{{\bm{W}}}
\def\mX{{\bm{X}}}

\DeclareMathAlphabet{\mathsfit}{\encodingdefault}{\sfdefault}{m}{sl}
\SetMathAlphabet{\mathsfit}{bold}{\encodingdefault}{\sfdefault}{bx}{n}

\newcommand{\R}{\mathbb{R}}

